\documentclass[journal]{IEEEtran}

\ifCLASSINFOpdf

\ifCLASSINFOpdf
   \usepackage[pdftex]{graphicx}
\else
\fi
\usepackage{amsmath}
\usepackage{amssymb}
\usepackage{xcolor}
\usepackage{color,soul}
\setul{0.5ex}{0.3ex}
\setulcolor{red}

\newcommand{\U}[1]{\underline{#1}}
\newcommand{\B}[1]{\textbf{#1}}
\newcommand{\I}[1]{\setulcolor{red}\ul{#1}}
\newcommand{\R}[1]{{#1}}

\begin{document}
%
\title{Forecasting future action sequences with attention: a new approach to weakly supervised action forecasting}
%
%
%

\author{{Yan Bin Ng,
        Basura Fernando}
\thanks{Institute of High Performance Computing, A*STAR, 1 Fusionopolis Way, 16-16, Connexis North Tower, Singapore 138632 e-mail: (see https://basurafernando.github.io/).}

\thanks{Manuscript published in IEEE Transactions on Image Processing.}}

%
%

\markboth{.}%
{Shell \MakeLowercase{\textit{et al.}}: Bare Demo of IEEEtran.cls for IEEE Journals}
%



\maketitle


%
\IEEEpeerreviewmaketitle

\begin{abstract}
	Future human action forecasting from partial observations of activities is an important problem in many practical applications such as assistive robotics, video surveillance and security. 
We present a method to forecast actions for the unseen future of the video using a neural machine translation technique that uses encoder-decoder architecture. 
The input to this model is the observed RGB video, and the objective is to forecast the correct future symbolic action sequence. 
Unlike prior methods that make action predictions for some unseen percentage of video one for each frame, we predict the complete action sequence that is required to accomplish the activity.
We coin this task action sequence forecasting.
To cater for two types of uncertainty in the future predictions, we propose a novel loss function. We show a combination of optimal transport and future uncertainty losses help to improve results. We evaluate our model in three challenging video datasets (Charades, MPII cooking and Breakfast). 

We extend our action sequence forecasting model to perform weakly supervised action forecasting on two challenging datasets, the Breakfast and the 50Salads.
Specifically, we propose a model to predict actions of future unseen frames without using frame level annotations during training.
Using Fisher vector features, our supervised model outperforms the state-of-the-art action forecasting model by 0.83\% and 7.09\% on the Breakfast and the 50Salads datasets respectively.
Our weakly supervised model is only 0.6\% behind the most recent state-of-the-art supervised model and obtains comparable results to other published fully supervised methods, and sometimes even outperforms them on the Breakfast dataset.
Most interestingly, our weakly supervised model outperforms prior models by 1.04\% leveraging on proposed weakly supervised architecture, and effective use of attention mechanism and loss functions.

\end{abstract}


\section{Introduction.}
\label{sec.intro}
\begin{figure}[t]
	\centering
	\includegraphics[width=1\columnwidth]{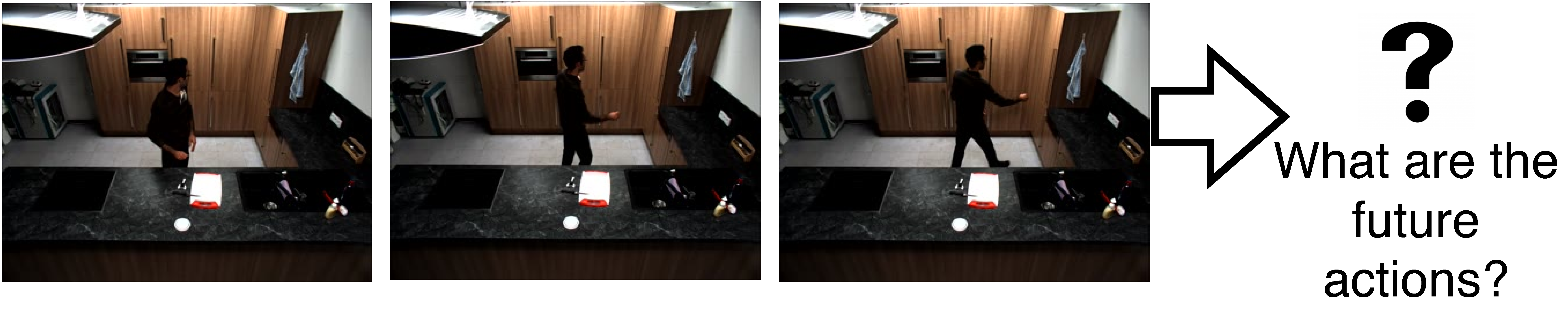}
	\caption{Someone is going towards the fridge. What are the plausible future sequence of actions? Our action sequence forecasting model predicts the future action sequence $\{$ open fridge $\succ$ take something $\succ$ close fridge $\}$, after processing the partial video. Our weakly supervised model predicts a label for each future frame without using any frame level annotations during training.}
	\label{fig.motivate1}
\end{figure}

We humans forecast others' actions by anticipating their behavior.
For example by looking at the video sequence in Fig.\ref{fig.motivate1}, we can say ``the person is going towards the fridge, then probably he will open the refrigerator and take something from it''. 
Our ability to forecast comes naturally to us.
We hypothesize that humans analyze visual information to predict plausible future actions, also known as mental time travel~\cite{Suddendorf2007}.
One theory suggests that humans' success in evolution is due to the ability to anticipate the future~\cite{Suddendorf2007}.
Perhaps, we correlate prior experiences and examples with the current scenario to perform mental time travel.

Recently, the human action prediction/forecasting problem has been extensively studied in the Computer Vision and AI community. 
The literature on this prediction topic can be categorized as early action~\cite{Kitani2012}, activity~\cite{AbuFarha2018,Lan2014}, and event prediction~\cite{Zeng2017}.
In early human action prediction, methods observe an ongoing human action and aim to predict \emph{the action in progress} as soon as possible~\cite{Kong2018} before it finishes.
This problem is also known as action anticipation in the literature~\cite{SadeghAliakbarian2017}.
As these methods predict an ongoing action before it finishes, they are useful for applications when future planning is not a major requirement.
In contrast, activity prediction aims at forecasting future action as soon as possible (not necessarily in the temporal order) and is useful in many robotic applications, e.g., human robot interaction. 
These methods can facilitate information for some level of future planning~\cite{Koppula2015}. In activity prediction, some methods observe $p$\% of the activity and then predict actions for $q\%$ of the future frames in the video.
Most interestingly, these methods predict actions per-frame which limits their practical application in many cases~\cite{AbuFarha2018}. 
The limitation of these methods are two fold. First, these methods need precise temporal annotations for each future frame during training. 
Even though this is feasible with small scale datasets, in practical applications obtaining labels for each frame is a challenging task. 
It is more feasible to obtain the sequence of actions without temporal extents. 
As an example, in Fig.\ref{fig.motivate1}, it is easy to obtain action sequence $<$ \texttt{open, take, close} $>$ rather than precise frame level annotations. In this paper we use only those coarse action sequence labels for training.
Secondly, most prior methods make the assumption about length of the video implicitly or explicitly~\cite{AbuFarha2018,Gammulle2019}. In contrast, our formulation does not make these rigid assumptions.

Alternatively, some methods observe $k$ number of actions in an activity and then predict only the next future action~\cite{Mahmud2017}.
However, we humans are able to forecast the future series of actions which allows us to plan for the future,  (e.g. if someone is going to cook a simple potato dish, probably we will see a sequence of actions such as peel $\succ$ cut $\succ$ wash $\succ$ boil). 
We humans are able to predict the future irrespective of video length or the number of frames. 
We aim to solve this challenging problem of forecasting future sequence of actions to complete an activity from the partial observations of the activity.
We call this task \emph{\textbf{action sequence forecasting}}.
This type of problems arise in practice, especially in robotics, e.g., robot assisted industrial maintenance, and assistive robotics in healthcare.

\begin{figure}[t!]
	\centering
	\includegraphics[width=\columnwidth]{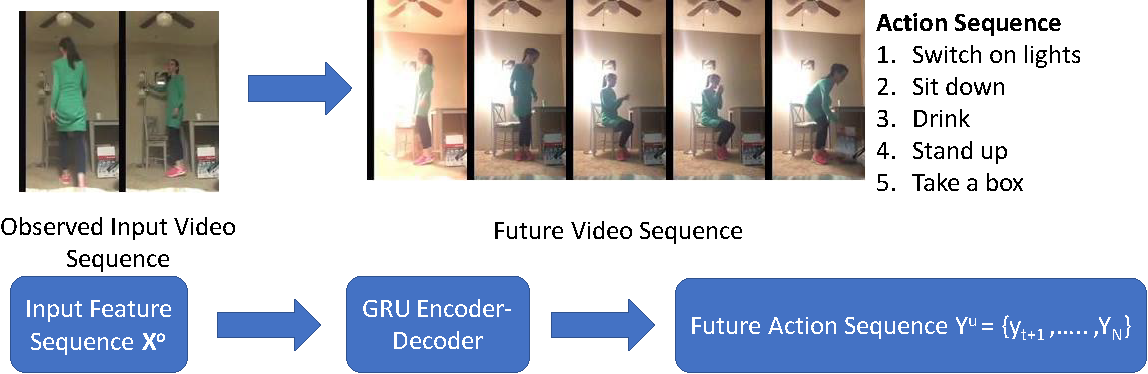}
	\caption{
		A high level illustration of our action sequence forecasting solution. 
		Given an input video, we train a GRU-based sequence-to-sequence machine translation model to forecast future action sequence. 
		Specifically, our method should know when to stop generating actions for the future. 
		In other words, we solve the problem of what steps (actions) are needed to finish the current activity of the person.
		During training we use the action sequences without temporal extents of each action. 
		At test time, our models are able to forecast future action sequence and a label for each future unseen frame.
	}
	\label{fig.illustrate1}
\end{figure}

Our model observes only a handful of actions within a long activity. 
Then it forecasts the sequence of actions for the future without making any assumptions on the length of video or using precise future temporal annotations.
In contrast to the majority of action anticipation and activity prediction models, ours is trained to predict \emph{the future action sequence} as shown in figure~\ref{fig.illustrate1}.
To solve this problem, there are several challenges that we need to tackle.
First, our method needs to implicitly infer the goal of the person.
Second, it should learn to what extent the person has completed the activity.
Finally, it has to infer what other actions are needed to accomplish the activity.
We formulate our solution such that all of this is learned in a data driven manner. 
Specifically, we make use of the complex relationship between observed video features and the future actions to learn a complex mapping  between them.
To facilitate that, we formulate it as a neural machine translation problem where the input is an observed RGB video and the target is a symbolic \emph{sequence of future actions}. 
Specifically, we use a recurrent encoder-decoder architecture.

Each future action depends on the past observed feature sequence and interestingly, some of the observed features are important in determining the future actions more than others.
For example, if our model predicts "adding sugar" as the future action, then it is more likely that our model gives a higher attention weight to frames having a cup or a mug.
Therefore, we make use of an attention mechanism that allows us to align-and-attend past features when generating future actions.
\R{
For each predicted future action, the attention mechanism processes the entire input feature sequence and selects the relevant set of observed frames that contain information about the future actions. 
Similar ideas have been explored before in other application domains, e.g., handwritten mathematical expression recognition~\cite{Zhang2017}.
Furthermore, our GRU encoder allows us to better model the temporal evolution of observed human actions and encode them into a temporally coherent hidden representation.
The attention mechanism query these hidden temporal representations and provide useful information to the decoder GRU to generate accurate future actions. 
}

Furthermore, the uncertainty of predictions increases with two factors; (1) the amount of data the model observes, and (2), how far into the future it predicts.
If our model observes more data, perhaps the predictions are likely to be reliable. 
Moreover, if the model predicts far into the future, then predictions are likely to be unreliable.  
We develop a novel loss function that allows us to consider these two factors and extend the traditional cross-entropy loss to cater for these uncertainties.
We also make use of optimal transport loss which allows us to tackle the exposure bias issue of this challenging sequence-to-sequence machine translation problem.
Exposure bias arises when we use cross-entropy loss to train neural machine translation models where it provides an individual action-level training loss (ignoring the sequential nature) which may not be suitable for our task.
The optimal transport loss is a more structured loss that aims to find a better matching of similar actions between two sequences, providing a way to promote semantic and contextual similarity between action sequences. In particular, this is important when forecasting future action sequences from observed temporal features.

Finally, we propose a model to predict action labels for future frames in a weakly supervised manner.
Weakly supervised action forecasting is useful for practical applications, specially when it is harder to obtain frame level annotations (or start, end of actions).
Specifically, we extend our action sequence forecasting model to perform action forecasting for future frames.
Our weakly supervised model generates \emph{pseudo representations} for future frames and then uses an attention mechanism to align them with forecasted future symbolic action sequence.
Using this mechanism it predicts labels for future unseen frames.
Our weakly supervised action forecasting method uses a novel GRU encoder-decoder architecture and we train this model using coarse action sequences (labels). 
Our model obtains results that are comparable to supervised methods and sometimes even outperforms them.

\R{
In a summary, our contributions are as follows:
%
\begin{itemize}
\item We propose an action sequence forecasting model that only utilizes the observed input frame sequence. Our architecture for weakly supervised action forecasting allows us to train with coarse annotations and predict action labels for frames at test time.
\item We propose new loss functions that handle the uncertainty in future action sequence forecasting and we demonstrate the usefulness of optimal transport and the uncertainty losses.
\item We extensively evaluate our method on four challenging action recognition benchmarks and obtain state of the art results for action forecasting. Our weakly supervised method obtains comparable results to prior supervised methods and in some datasets even outperforms them.
\end{itemize}
}

\section{Related work.}
\label{sec.related}
We categorize the related work into three, 1. early action prediction and anticipation, 2. activity prediction and 3. weakly supervised action understanding and 4. machine translation.\\

\noindent
\textbf{Early action prediction and anticipation:}
Early action prediction aims at classifying the action as early as possible from partially observed action video.
Typically, experiments are conducted on well segmented videos containing a single human action. 
In most prior work, methods observe about 50\% of the video and then predict the action label~\cite{Ryoo2011,Kong2014,SadeghAliakbarian2017}.
In particular these methods can be categorized into four types.
Firstly, some generate features for the future and then use classifiers to predict actions using generated features~\cite{Shi2018,Vondrick2016}. 
Feature generation for future action sequences containing a large number of actions is a challenging task and therefore, not feasible in our case.
Secondly, \cite{SadeghAliakbarian2017,Jain2016,Ma2016} develop novel loss functions to cater for uncertainty in the future predictions.
Our work also borrows some concepts from these methods to develop loss functions but ours is applied over the future action sequence in contrast to applying over an action.
Thirdly, some anticipation methods generate future RGB images and then classify them into human actions using convolution neural networks ~\cite{Zeng2017a,Wang2017}. 
However, generation of RGB images for the future is a very challenging task specially for longer action sequences.
Similarly, some methods aim to generate future motion images and then try to predict  action for the future~\cite{Rodriguez2018}. 
However, we aim to forecast action sequences for unseen parts of the human activity and are more challenging than action anticipation.
Therefore, action anticipation methods can not be used to solve our problem.
Furthermore, our approach uses coarse level video annotations during training which allows someone to scale up our method for very large scale problems as-well-as for domains where obtaining frame level annotations is difficult.

\noindent
\textbf{Activity prediction:}
Some action forecasting methods assume that the number of future frames is given and predict the action label for each future frame~\cite{AbuFarha2018,Gammulle2019,ke2019time}. 
These methods (\cite{AbuFarha2018,Gammulle2019,ke2019time}) require precise temporal annotations at frame-level during training and some methods use ground truth labels for observed actions~\cite{Gammulle2019} during inference.
Several action forecasting methods are evaluated in~\cite{AbuFarha2018} including  CNN and RNN based approach.
\R{In particular this method~\cite{AbuFarha2018} uses frame level annotations of observed data to  train a frame classification model. This model predicts actions of the observed frames. Then using the generated actions of the observed frames, it directly predicts the actions for the future frames. Furthermore, the CNN, RNN methods of~\cite{AbuFarha2018} predicts future actions for frames in a segment-wise manner until the desired prediction level is reached. 
In contrast, our model does not make any assumptions about the length of the sequence and rely on the model to emit the end-of-sequence token when predicting future action sequences. Furthermore, when predicting action labels for the future frames in a weakly supervised manner, our method generates pseudo representations for future frames and predicts action labels per-frame using the attention mechanism.} 
Method in~\cite{Gammulle2019} uses a memory network which is also an encoder-decoder method.
However, our model architecture and the loss functions used are different from~\cite{Gammulle2019}.
\R{Our model does not use frame level annotations and use only coarse video level annotations during training. In contrast, the method in~\cite{Gammulle2019} uses frame level annotations and frame features during training and even frame level annotations of observed frames during testing. 
First, both the observed frame sequence and the action sequence are encoded with a LSTM unit and then they are concatenated and fed to a memory network decoder to generate future actions.
In contrast, our weakly supervised action forecasting model relies on a new encoder-decoder architecture with three dedicated decoders to generate future action labels for frames in a weakly supervised manner.
We use only coarse action labels during training and our model does not need any annotated frames during inference as in~\cite{Gammulle2019}.}
Model presented in~\cite{ke2019time} consists of two parts, i.e., temporal feature attention module, and time-conditioned skip connection module for action forecasting.
Our model is different from all these methods due to the differences in model architecture and type of supervision used.
We aim to predict the future sequence of action (e.g. wash $\succ$ clean $\succ$ peel $\succ$ cut) and assign a label for each future frame in a weakly supervised manner without using frame level annotations during training.

Some activity prediction methods aim at predicting the next action in the sequence~\cite{Mahmud2017,Qi2017} or focus on first person human actions~\cite{Rhinehart2017,Bokhari2016}.
Specifically, \cite{Mahmud2017} used to predict the next action using the previous three actions using motion, appearance, and object features with a two layered stacked LSTM. 
Authors in~\cite{Qi2017} use stochastic grammar to predict the next action in the video sequence. 
Even though these methods can be extended to predict the sequence of actions by recursively applying them, we face two challenges. 
Firstly, errors may propagate making future actions more wrong, and secondly it may not know when to stop producing action symbols, which is important when the actions are part of some larger activity. 
Sequence-to-sequence machine translations are naturally able to address both these two issues~\cite{Sutskever2014}.
We make use of this strategy to solve our problem.

\noindent
\textbf{Weakly supervised action understanding:}
To the best of our knowledge we are the first to present a model for weakly supervised action forecasting.
Weakly supervised methods are used for tasks such as action detection~\cite{Nguyen_2018_CVPR,Fernando2020} and segmentation~\cite{huang2016connectionist}.
Authors in~\cite{huang2016connectionist} utilize weak annotations for action classification by aligning each frame with a label in a recurrent network framework. 
Specifically, they use connectionist temporal classification architecture with dynamic programming to align actions with frames.
Authors in~\cite{Singh2017} learn to find relevant parts of an object/action after randomly suppressing random parts of  images/videos. 
This method focuses only on very few discriminative segments. 
Authors in~\cite{Shou2018} directly predict the action boundaries using outer-inner-contrastive loss in a weakly supervised manner. 
Authors in~\cite{Nguyen_2018_CVPR} propose two terms that minimize the video-level action classification error and enforce the sparsity when selecting segments for action detection.
Authors in~\cite{Paul2018} use attention-based mechanism to identify relevant frames and apply a pairwise video similarity constraint in a multiple instance framework. 
Our weakly supervised model is different from what has been used in prior work with respect to two reasons.
First, our model generates features for unseen frames and then assigns a label for each using only coarse sequence annotations as shown in Fig~\ref{fig.illustrate1}.
Secondly, our network architecture contains a hierarchy of GRU encoders and decoders. 
To be precise, it consists of a single encoder and three decoders dedicated to predict and assign action labels for future frames.

\noindent
\textbf{Machine translation.}
Our method is also related to machine translation methods~\cite{Sutskever2014,Bahdanau2014,Venugopalan2015,Yu2016}.
However, none of these works use machine translation for action sequence forecasting from videos.
Typically, machine translation is used for language tasks~\cite{Sutskever2014,Bahdanau2014}.
To the best of our knowledge, we are the first to use neural machine translation for translating a sequence of RGB frames (a video) to a sequence of future action labels with \emph{weak supervision}.
Indeed, machine translation has been used for unsupervised learning of visual features~\cite{Srivastava2015} in prior work which is related to us.
But they did not use it for predicting future action sequences.

\section{Future action sequence generation.}
\label{sec.method}
\subsection{Problem}
We are given a video in which a human is performing an activity.
Our model only observes the initial part of the video containing initial sequence of actions.
The objective of this work is to train a model to predict the future unseen sequence of actions.
A visual illustration of this model is shown in figure~\ref{fig.illustrate1}.
Let us denote the observed RGB video by $X^o=\left< x^o_1, x^o_2, x^o_3, \cdots, x^o_p \right >$ where $x^o_p$ is the $p^{th}$ frame. 
The observed action sequence is denoted by $Y^o=\left< y^o_1, y^o_2, \cdots y^o_{\mathcal{P}} \right>$ (note that $p \neq \mathcal{P}$) and the future unseen ground truth action sequence by $Y^u=\left< y^u_1, y^u_2, \cdots y^u_{\mathcal{N}} \right>$ where each action $y \in \mathcal{Y}$ and $\mathcal{Y}$ is the set of action classes and the start time of each action $y_i$ is before or equal to the start time of $y_{i+1}$.
In contrast to most other action forecasting problems, we do not use frame level annotations. 
Each observed ($y^o_t$) or future unseen action ($y^u_t$) may span over multiple frames and we do not explicitly use the start and end time of each action.
However, during inference we are able to predict an action label for future unseen frames or explicitly infer the start and end of each future action.

First, we present our default future action sequence forecasting model in section~\ref{sec.future.act.seq.gen}. 
Second, we extend this model to infer start and end of future actions only using coarse labels sequences $Y^o$ and $Y^u$ for training.
This weakly supervised action forecasting method is presented in section~\ref{sec:wsaf}.
Our fully supervised model which uses frame level annotations for observed and future actions is presented in ~\ref{sec:saf}.

\subsection{Future action sequence forecasting model} 
\label{sec.future.act.seq.gen}
In contrast to other action forecasting methods that operates at frame level (or clip level), we do not know the label of each observed RGB frame $ x^o_p$. 
Our model has access to frame sequence $X^o$ only.
We train a model $\phi(,\Theta)$ that predicts unseen action sequence $Y^u$ from seen RGB feature sequence $X^o$ where $\Theta$ are the parameters of model, i.e. $Y^u = \phi(X^o,\Theta)$.
We do not make use of ground-truth action sequence $Y^o$ during training or inference.
Therefore, our method does not need any frame level action annotations as in prior action forecasting methods~~\cite{AbuFarha2018,Gammulle2019}.

We formulate this problem as a sequence-to-sequence machine translation problem~~\cite{Sutskever2014,Bahdanau2014,Venugopalan2015,Yu2016} where we use observed rgb sequence $X^o$ as the input sequence. Then the symbolic unseen action sequence $Y^u$ is the target sequence. Specifically, we use an GRU-based encoder-decoder architecture. 
Our hypothesis is that the encoder-decoder machine translation would be able to learn the complex relationship between seen feature sequence and future actions.
To further improve the model predictive capacity, we also use attention over encoder hidden state when generating  action symbols for the future and use novel loss functions to tackle uncertainty. 
Next we describe our model in detail.

\subsubsection{GRU-encoder-decoder}
\label{sec:GRU}
We use GRU-based encoder-decoder architecture for translating video sequence into future action sequence.
Our encoder consists of a bi-directional GRU cell. 
Let us define the encoder GRU cell which takes the observed feature sequence consisting of $p$ elements as input. 
We define the encoder GRU by $f_e()$ for time step $t$ as follows:
\begin{equation}
 \overrightarrow{\mathbf{h}}_t, \overleftarrow{\mathbf{h}}_t = f_e(\mathbf{x^o_t}, \overrightarrow{\mathbf{h}}_{t-1}, \overleftarrow{\mathbf{h}}_{t-1})
\end{equation}
where $\overrightarrow{\mathbf{h}}_t, \overleftarrow{\mathbf{h}}_t \in R^{\mathcal{D}}$ are the forward and backward hidden states at time $t$.
The initial hidden state of the encoder GRU is set to zero.
Then we make use of a linear mapping $W_{e} \in R^{2\mathcal{D} \times 
\mathcal{D}}$ to generate a unified representation of both forward and backward 
hidden states for each time step $\overrightarrow{\mathbf{h}}_t, 
\overleftarrow{\mathbf{h}}_t$ as follows:
\begin{equation}
 \mathbf{h_t} = [\overrightarrow{\mathbf{h}}_{t-1}, \overleftarrow{\mathbf{h}}_{t-1}]\times W_{e}
\end{equation}
where $[\cdot,\cdot]$ indicates the concatenation of forward and backward hidden states.
Therefore, the outcome of the encoder GRU is a sequence of hidden state vectors denoted by $\mathbf{H} = \left< \mathbf{h^o_1}, \mathbf{h^o_2}, \cdots \mathbf{h^o_p} \right>$.
The bi-directional GRU encode more contextual information which might inherently enable the model to infer the intention of person doing the activity.
The decoder is a forward directional GRU $f_d()$, that generates the decoder hidden state $\mathbf{g_q} \in R^{\mathcal{D}}$ at decoding time step $q$ define as follows:
\begin{equation}
 \mathbf{g_q} = f_d([\mathbf{c_{q-1}},\mathbf{\hat{y}_{q-1}}], \mathbf{g_{q-1}})
 \label{eq.decoder}
\end{equation}
where $\mathbf{\hat{y}_{q-1}}$ is the predicted target action class score vector at step q-1.
The input to decoder GRU $f_d()$ at time step is a concatenation of the context vector $\mathbf{c_{q-1}}$ and the previously predicted action score vector $\mathbf{\hat{y}_{q-1}}$ denoted by $[\mathbf{c_{q-1}},\mathbf{\hat{y}_{q-1}}]$.
We obtain the action score vector at step $q$ of the decoder using following linear mapping:
\begin{equation}
 \mathbf{\hat{y}_{q}} = \mathbf{g_q} \times U
\end{equation}
where $U \in R^{\mathcal{D} \times |\mathcal{Y}|}$ is a learnable parameter. Note that the output symbol at step $q$ of the decoder is obtain by $\texttt{argmax}$ operator, i.e., 
$\hat{y}_q = \texttt{argmax}~~\mathbf{\hat{y}_{q}}$. 
The decoder is initialized by the final hidden state of the encoder (i.e. $\mathbf{g_0}=\mathbf{h^o_p}$ where $\mathbf{h^o_p}$ is the final hidden state of the encoder).
The initial symbol of the decoder is set to SOS (start of sequence symbol) during training and testing.
The decision to include the previous predicted action $\mathbf{\hat{y}_{q-1}}$ as an input in the decoder is significant as now the decoder model has more semantic information during the decoding process. One choice would be to simply ignore the previously predicted action symbol. However, that would hinder the predictive capacity of the decoder as decoder is not explicitly aware of what it produced in the previous time step. Conceptually, now the decoder is trying to find the most likely next symbol $P(y_q|y_{q-1},\mathbf{g_{q-1}},\mathbf{c_{q-1}} )$ using both previous symbol and the contextual information.

Next we describe how to generate the context vector $\mathbf{c_{q-1}}$ which summarizes the encoder-decoder hidden states using attention mechanism.

\subsubsection{Attention over encoder hidden state} 
\label{sec:attention}
It is intuitive to think that not all input features contribute equally to generate the output action symbol $\hat{y}_q$ at decoder step $q$.
Therefore, we propose to make use of attention over encoder hidden states $\mathbf{H}$ to generate the context vector $\mathbf{c_{q-1}}$ which serves as an part of input to the decoder GRU.
Specifically, to generate $\mathbf{c_{q-1}}$, we linearly weight the encoder hidden vectors $\mathbf{H} = \left< \mathbf{h^o_1}, \mathbf{h^o_2}, \cdots \mathbf{h^o_p} \right>$, i.e., 
\begin{equation}
 \mathbf{c_{q}} = \sum_i \frac{exp(\alpha_i^q)}{\sum_j exp(\alpha_j^q) }  \mathbf{h^o_i}
\end{equation}
where $\alpha_i^q$ is the weight associated with the encoder hidden state $\mathbf{h^o_i}$ to obtain $q$-th context vector defined by the following equation.
\begin{equation}
 \alpha_i^q = tanh([\mathbf{h^o_i}; \mathbf{g_q}] \times W_{att}) \times V
\end{equation}
Here $W_{att} \in R^{2\mathcal{D} \times \mathcal{D}}$ and $V \in \mathcal{D}$ 
are learnable parameters and $\alpha_i^q$ depends on how well the 
encoder-decoder hidden states $\mathbf{h^o_i}, \mathbf{g_q}$ are related.
This strategy allows us to attend all encoder hidden states $\mathbf{H} = \left< \mathbf{h^o_1}, \mathbf{h^o_2}, \cdots \mathbf{h^o_p} \right>$ when generating the next action symbol using decoder GRU.
During training we make use of the teacher forcing strategy to learn the model 
parameters of the encoder-decoder GRUs where we randomly choose to use $\mathbf{y_q}$ instead of 
$\mathbf{\hat{y}_{q}}$ in equation~\ref{eq.decoder} with a probability of 0.5.
This is to make sure that the inference strategy is not too far away from the 
training strategy and that convergence takes place faster. During inference, given the input features sequence, we pass 
it thorough the encoder-decoder to generate future action sequence until we hit 
the end-of-sequence symbol (EOS). The model is also trained with 
start-of-sequence (SOS) symbol and EOS.

\subsubsection{Tackling the uncertainty}
\label{sec:losses}
Correctly predicting the future action sequence from a partial video is challenging as there are more than one plausible future action sequences. 
This uncertainty increases with respect to two factors; 1. to what extent we have observed the activity, (the more we observe, the more information we have to make future predictions) and 2. how far into the future we are going to predict using observed data (if we predict too far into the future, there are more possibilities and more uncertainty). 
To tackle these two factors, we propose to modify the cross-entropy loss which is typically used in sequence-to-sequence machine translation\footnote{This strategy may be applicable to other loss functions as well.}.
Let us assume that we have observed $\mathcal{P}$ number of actions, and we are predicting a total of $\mathcal{N}$ number of action symbols. 
Let us denote the cross-entropy loss between the prediction ($\mathbf{\hat{y}_{q}}$) and the ground truth ($\mathbf{y}_{q}$) by $\mathcal{L}(\mathbf{\hat{y}_{q}},\mathbf{y}_{q})$. Then our novel loss function that handles the uncertainty ($L_{un}(\hat{Y}^{u},Y^u)$) for a given video $X^o,Y^u$ is define by
\begin{equation}
 L_{un} = (1-exp(-\frac{\mathcal{P}}{\mathcal{N}})) \sum_{q=1}^{\mathcal{N}} exp(-q) \mathcal{L}(\mathbf{\hat{y}_{q}},\mathbf{y}_{q}) 
 \label{eq.un}
\end{equation}
where the term $(1- exp(-\mathcal{P}/\mathcal{N}))$ takes care of shorter observations and makes sure that longer action observations contributes more to the loss function. If the observed video contains less actions (information), then predictions made by those are not reliable and therefore does not contribute much to the overall loss.
Similarly, the second inner term $ exp(-q) \mathcal{L}(\mathbf{\hat{y}_{q}},\mathbf{y}_{q})$ makes sure that those predictions too far into the future make only a small contribution to the loss.
If our model makes a near future prediction, then possibly model should do a better job and if it makes an error, we should penalize more.
During training, we make use of sequential data augmentation to better exploit the above loss function.
In-fact, for given a training video consist of $\mathcal{M}$ actions (i.e. $Y=\left< y_1, y_2, \cdots y_\mathcal{M} \right>$), we augment the video to generate $\mathcal{M}-1$ observed sequences where $Y^o=\left< y^o_1, y^o_2, \cdots y^o_t \right>$ and $Y^u=\left< y^o_{t+1}, \cdots y^o_M \right>$ for $t=\{1,\cdots,{M-1}\}$. Then we train our networks with these augmented video sequences with the uncertainty loss.

\subsubsection{Optimal Transport Loss (OT)}
\label{sec:ot.algo}

\R{
So far, the cross-entropy loss is applied over actions in a point-wise manner without taking into account the topological or the geometric structure of action space.
The element-wise cross-entropy loss obtained for action at step-q of the decoder only relies on the ground-truth action at step-q and it does not take the sequence-to-sequence structural nature of the task.
Moreover, when the predicted sequence is longer than the ground truth, the cross-entropy loss requires adhoc end-of-sequence token (class) to handle this.
However, ideally, the encoder-decoder model should be able to predict the target action sequence of the future by considering structural nature of this task.}

The optimal transport defines a distance measure between probability distributions over a metric space by considering the topology, and in our case the topology of actions sequences. 
It is desirable to exploit this metric structure in the action sequence space using optimal transport loss~\cite{Peyre2019} over predicted action sequences.
We propose to make use of optimal transport loss of~\cite{Peyre2019} defined by 
\begin{equation}
 D_c(\mathcal{\mu},\mathcal{\nu}) = \inf_{\gamma \in \Pi(\mathcal{\mu},\mathcal{\nu})} \mathbb{E}_{(x,y)\sim \gamma}[c(\mathbf{x,y})] 
\end{equation}
where $\Pi(\mathcal{\mu},\mathcal{\nu})$ is the set of all joint distributions $\gamma(\mathbf{x},\mathbf{y})$ with marginals $\mathcal{\mu}(\mathbf{x})$ and $\mathcal{\nu}(\mathbf{y})$ and $c(\mathbf{x},\mathbf{y})$ is the cost function for moving $\mathbf{x}$ to $\mathbf{y}$ in the sequence space. We take the cost function to be the L2 norm i.e. $c(\mathbf{x,y}) = \|\mathbf{x}-\mathbf{y}\|_2$. 

Specifically, we consider the optimal transport distance between two discrete action distributions $\mu,\nu \in \mathbf{P}(\mathbb{A})$ of the action sequences where $\mathbb{A}$ is the action space. 
The discrete distributions $\mu,\nu$ can be written as weighted sums of Dirac delta functions i.e. $\mu = \sum_{i=1}^n \mathbf{u}_i\delta_{\mathbf{x}_i}$ and $\nu = \sum_{j=1}^m \mathbf{v}_j\delta_{\mathbf{y}_j}$ with $\sum_{i=1}^n \mathbf{u}_i = \sum_{j=1}^m \mathbf{v}_j = 1$. Given a cost matrix $\mathbf{C} \in  \mathbb{R}_+^{n \times m}$ where $\mathbf{C}_{ij}$ is the cost from $\mathbf{x}_i$ to $\mathbf{y}_j$,  the optimal transport loss is equivalent to
\begin{equation}
 L_{ot}(\mu,\nu) = \min_{\mathbf{P} \in \Pi(\mathbf{u},\mathbf{v})} \sum_{i,j} \mathbf{P}_{ij}\mathbf{C}_{ij}
 \label{eq.ot}
\end{equation}
where $\Pi(\mathbf{u},\mathbf{v}) = \{\mathbf{P} \in \mathbb{R}_+^{n \times m}\} | \mathbf{P}\mathbf{1}_m = \mathbf{u}, \mathbf{P}^\top\mathbf{1}_n = \mathbf{v} \}$ and $\mathbf{1}_n$ is a n-dimensional vector of all ones.

\R{
The minimum $P^*$ in equation~\ref{eq.ot} is the ideal optimal transport solution that caters for the topological structure of predicted and ground truth action sequences.
The cost function is defined in the Euclidean space over action predictions as follows:
\begin{equation}
C_{ij} = ||\hat{\mathbf{s_i}} - \mathbf{s_j}||_2 
\end{equation}
where $\hat{\mathbf{s_i}}$ is the predicted action score vector at step $i$ and $\mathbf{s_j}$ is the one-hot-vector obtained from the ground truth action sequence $Y^u$ at step j.
Note that both $\hat{Y}^{u}$ and $Y^u$ are discrete action symbol sequences and the optimal transport loss is complementary to the cross entropy loss and vice-versa.}

Because the optimal transport assignment problem is formulated as a permutation problem, we make use of the Sinkhorn divergence method proposed in~\cite{Feydy2019} to estimate the optimal transport loss in equation~\ref{eq.ot}.
Using this Sinkhorn algorithm implementation proposed in~\cite{Feydy2019}, we compute the optimal transport loss between the predicted and ground-truth action probability distributions from equation~\ref{eq.ot}.
Let $S$ be the $\mathcal{N} \times |\mathcal{Y}|$ ground truth tensor where $S_{jk}$ contains the probability value of action $k$ in step $j$ (i.e. each row $j$ is equal to one-hot vector $\mathbf{s_j}$), and $\hat{S}$ be the corresponding tensor containing the predicted probability values (i.e. each row $i$ is equal to probability vector $\hat{\mathbf{s_i}}$). 
The optimal transport loss ($L_{ot}$) computed using the Sinkhorn algorithm is then denoted by $L_{sh}(\hat{S},S)$.
The combination of both losses is given by the following:
\begin{equation}
 L_{total} = L_{un}(\hat{Y}^{u},Y^u) + \beta \times L_{sh}(\hat{S},S),
 \label{eq.otandun}
\end{equation}
where $\beta$ is the trade-off parameter and $L_{un}(\hat{Y}^{u},Y^u)$ is obtained by equation~\ref{eq.un}.

\section{Weakly supervised future action forecasting}
\label{sec:wsaf}
\begin{figure}
	\includegraphics[width=.99\columnwidth]{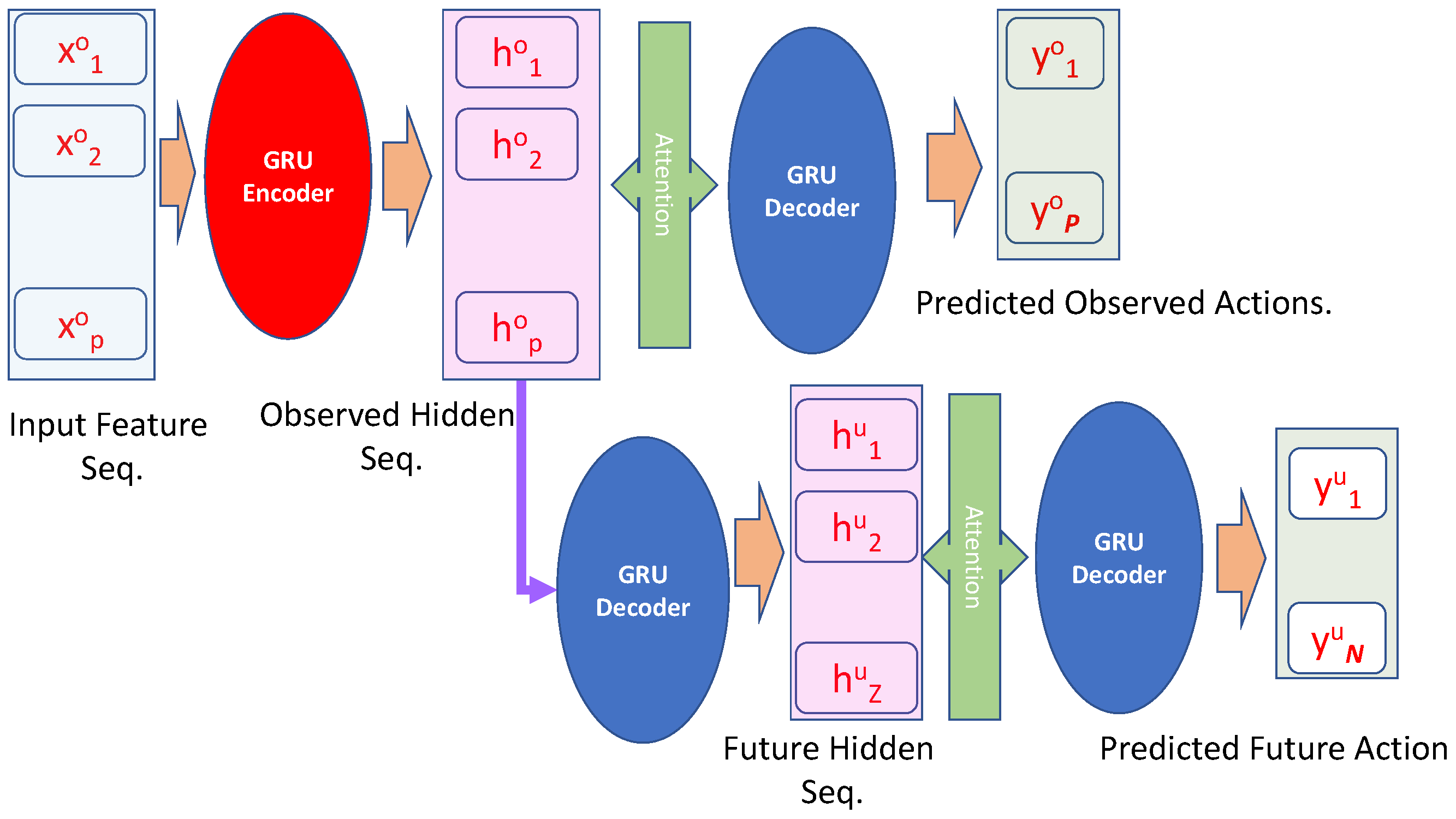}
	\caption{A visual illustration of our weakly supervised action forecasting architecture. 
		It consists of a encoder and three decoder GRUs. 
		The first decoder uses attention mechanism to align observed hidden sequence $H^o$ with the observed action sequence $Y^o$. 
		The second GRU decoder processes last observed hidden state $h^o_p$ and then decodes to generate pseudo states for the future features denoted by $H^u = \left< h^u_1, \cdots, h^u_Z \right>$. 
		The attention mechanism is used to align predicted future action sequence $Y^u$ with pseudo hidden states $H^u$. This allows us to estimate a label for each future hidden state using the attention weights and action scores $< \mathbf{y^u_1, \cdots. y^u_N} >$.
	}
\label{fig.weakarchi}
\end{figure}
In this section we present a method to infer the temporal extent of each future action using only the coarse labels of observed and unseen future action sequences.
Therefore, essentially our method is a weakly supervised action forecasting method.
Specifically, during training we make use of observed feature sequence $X^o=\left< x^o_1, x^o_2, x^o_3, \cdots, x^o_p \right >$, observed action sequence $Y^o=\left< y^o_1, y^o_2, \cdots y^o_{\mathcal{P}} \right>$ and the future unseen action sequence $Y^u=\left< y^u_1, y^u_2, \cdots y^u_{\mathcal{N}} \right>$.
Note that both $Y^o$ and $Y^u$ are coarse sequences, i.e. we do not know the temporal extent of each action.
Both $Y^o$ and $Y^u$ are used to compute the loss during training.
At test time, we predict future action sequence $\hat{Y^u}$ and a label of for each unseen frame only using observed feature sequence $X^o$. To do that, we use the attention mechanism presented in section~\ref{sec:attention}.
High level illustration of our novel architecture is shown in Fig.~\ref{fig.weakarchi}. Now we give more details of our weakly supervised action forecasting method

First, our model processes the feature sequence $X^o$ using a GRU encoder $f^o_e()$ to obtain a hidden state sequence $H^o$ of the same length as $X^o$.
\begin{equation}
	H^o = f^o_e(X^o)
\end{equation}
Then, a GRU decoder $f^o_d()$ with attention is used to decode $H^o$ to obtain the observed action sequence $\hat{Y^o}$.
The encoder-decoder with attention used here (i.e. $f^o_e()$ and $f^o_d()$) is explained in section~\ref{sec:GRU} and ~\ref{sec:attention}.

The second GRU decoder $f^{u}_{d}()$ \textbf{without attention} takes the last hidden state ${h^o_p}$ of $H^o$ as the initial hidden state and decodes to generate a sequence of future hidden states for unseen frames.
Let us assume there are $Z$ number of future unseen frames. 
Therefore, $f^{u}_{d}()$ generates the future hidden state sequence $H^u=\left< h^u_1, \cdots {h^u_Z} \right >$ as follows:
\begin{equation}
H^u = f^{u}_{d}(h^o_p).
\end{equation}
We call these future unseen hidden vectors as \textbf{pseudo states} and in practice $Z$ is known or given to us.
The final GRU decoder ($f^{uy}_{d}$) with attention is applied over these pseudo state sequence $H^u$ to obtain future action sequence $\hat{Y^u}$ as follows:
\begin{equation}
\hat{Y^u} = f^{uy}_{d}(H^u).
\end{equation}
The attention allows us to align each future hidden state $h^u_t$ with the corresponding future label $\hat{y^u_q}$ of $\hat{Y^u}$.
Similar to equation~\ref{eq.decoder}, inputs to this decoder are the predicted future action score and the context vectors.
Let $\alpha_t^q$ be the attention score on $t^{th}$ unseen hidden state $h^u_t$ for generating $q^{th}$ action symbol $\hat{y^u_q}$ of $\hat{Y^u}$.
The attention scores are obtained as explained in section~\ref{sec:attention}.
The attention score indicates the contribution of future frame $x^u_t$ for generating action $y^u_q$ of the future.
Therefore, we can assign action score $s_t$ for each future frame $x^u_t$ using the following equation
\begin{equation}
	s_t =\sum_q \alpha_t^q \times \mathbf{y^u_q}
\end{equation}
where $\mathbf{y^u_q}$ is the score vector of $q^{th}$ future action symbol of $\hat{Y^u}$. 
Even though, we have never observed any of future features, this formulation allows us to generate pseudo hidden states for future and then decode that to generate future action sequence. 
This attention mechanism allows us to assign a label for each future frame without using any explicit frame level annotation during training.

To train $f^o_e(), f^o_d(), f^u_d(), h^{uy}_{d}()$ we use combination of two losses as follows:
\begin{equation}
Loss =  \mathcal{L}(Y^o,\hat{Y^o}) + \gamma \mathcal{L}(Y^u,\hat{Y^u})
\label{eq.loss.joint}
\end{equation}
where the loss functions $L()$ are explained in section~\ref{sec:losses}.
As before, we use end-of-sequence token (EOS) during training and testing for both $\hat{Y^o}$ and $\hat{Y^u}$.
In our implementation we make sure that the feature dimensions of $X,H^o,H^u$ are the same.

\subsection{Fully supervised action forecasting}
\label{sec:saf}
We also extend the method presented in section~\ref{sec:wsaf} for fully supervised action forecasting.
In this case, the observed and future action sequences $Y^o, Y^u$ are frame specific and the loss function in Eq.~\ref{eq.loss.joint} is applied over frame level action annotations during training. 
However, at inference, model takes observed feature sequence $X^o$ as input and predicts action labels for each future frame.
We do not use EOS token for fully supervised model.

\section{Experiments.}
\label{sec.exp}
In this section we extensively evaluate our model using three challenging action recognition datasets, namely the Charades\cite{Sigurdsson2016},  MPII Cooking\cite{Rohrbach2012} and Breakfast\cite{Kuehne2014} datasets. Next we give a brief introduction to these 
datasets.
\begin{figure*}[t]
\centering
\includegraphics[width=1.4\columnwidth]{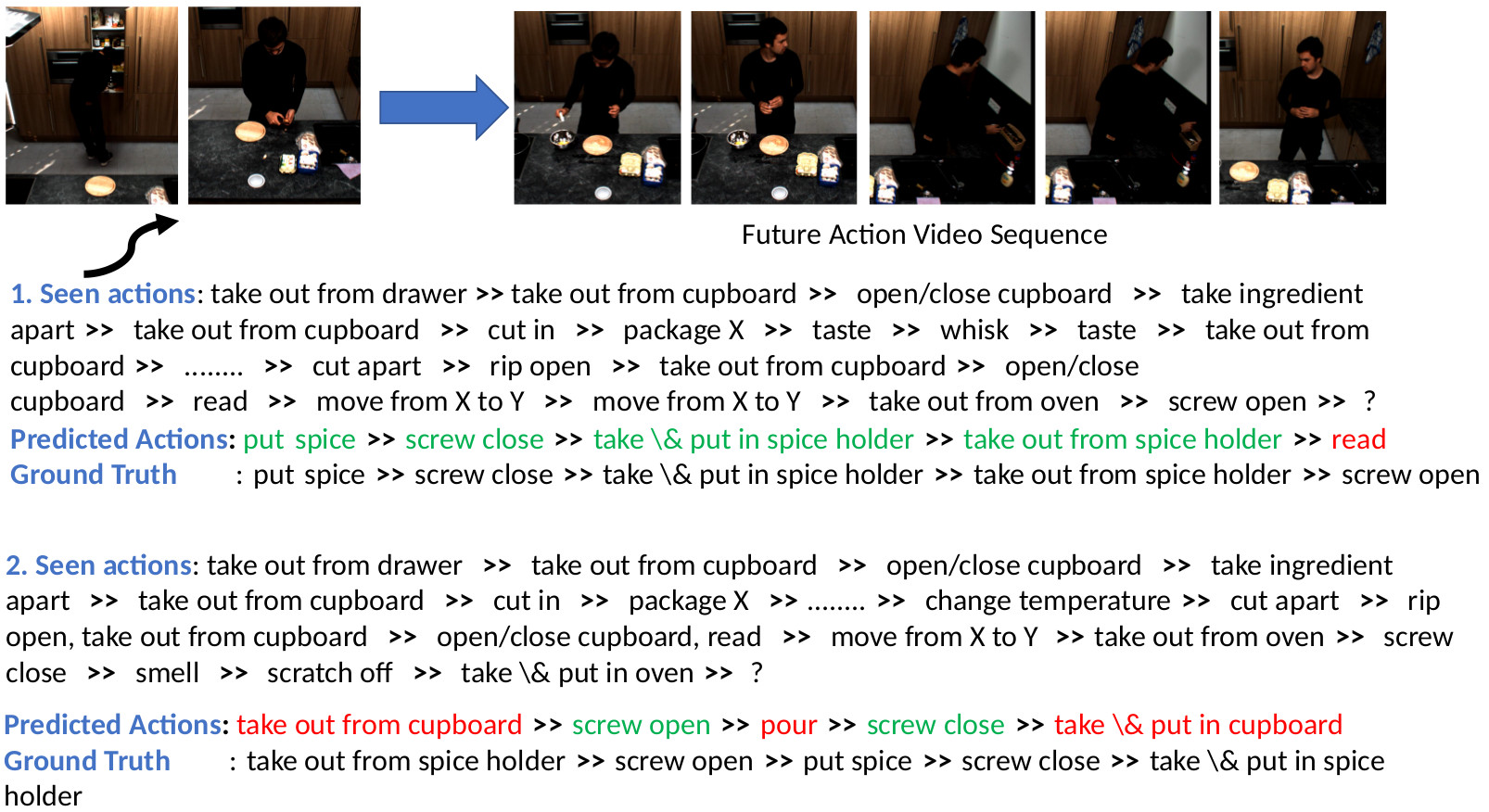}
\caption{Qualitative results obtained with our method on MPII Cooking dataset. Correctly predicted actions are shown in green and the wrong ones in red.}
\label{fig.results1}
\end{figure*}
MPII-Cooking Dataset has 65 fine grained actions, 44 long videos with a the total length of more than 8 hours. 
Twelve participants interact with different tools, ingredients and containers to make a cooking recipe.  
We use the standard evaluation splits where total of five subjects are permanently used in the training set. 
Rest six of seven subjects are added to the training set and all models are tested on a single subject and repeat seven times in a seven fold cross-validation manner. In this dataset, there are 46 actions per video on average.

Charades dataset has 7,985 video for training and 1,863 videos for testing.  
The dataset is collected in 15 types of indoor scenes, involves interactions with 46 object classes and has a vocabulary of 30 verbs leading to 157 action classes\cite{Sigurdsson2016}. 
On average there are  6.8 actions per video which is much higher than other datasets having more than 1,000 videos.

Breakfast dataset\cite{Kuehne2014} consist of 1,712 video where 52 actors making breakfast dishes. There are 48 fine-grained actions classes and four splits. On average each video consists of 6.8 actions per video.

There are no overlapping actions in Breakfast and Cooking datasets.
Charades has handful of videos with overlapping actions.
To generate ground truth action sequences, we sort the list of actions by  start times, ignoring the end time of the actions.

\subsection{Performance evaluation measures:}
We measure the quality of generated future action sequences using BLEU-1 and BLEU-2 scores\cite{Papineni2002}. 
These are commonly used in other sequence evaluation tasks such as image captioning.
We use the standard BLEU score definition proposed in the Machine Translation and Natural Language Community\cite{Papineni2002} which is also publicly implemented in Python nltk toolbox.
We also report sequence-item classification accuracy which counts how many times the predicted sequence elements match the ground truth in the exact position. 
Furthermore, we also report the mean average precision (mAP) which does not account for the order of actions. 
To calculate mAP, we accumulate the action prediction scores of the unseen video and compare it with the ground truth.
BLEU-1, BLEU-2 and sequence-item classification accuracy reflects the sequence forecasting performance while the mAP only accounts for holistic future action classification performance discarding the temporal order of actions.

The use of BLEU scores is somewhat novel in action forecasting.
In machine translation, BLEU score is used to compare a candidate sequence of words against a reference translation. Similarly, we use BLEU scores in the context of action sequence forecasting to provide a precision measure over action sequences.
For example, BLEU-2 score indicates the precision of each models' ability to correctly predicts two-action compositions (e.g open $\succ$ close, wash $\succ$ peel).
Therefore, BLEU scores provides complementary information to sequence-item classification accuracy.

\subsection{Feature extraction and implementation details:}
Unless specifically mentioned, we use effective I3D features \cite{Carreira2017} as the video representation for all datasets. 
First, we fine-tune I3D networks for video action classification using provided video level annotations. 
Afterwards, we extract 1024-dimensional features to obtain a feature sequence for each video.

\subsection{Evaluating our model}
\R{
In the following sections we evaluate various aspects of our model aiming to provide some insights to the reader.
First, we evaluate our action sequence forecasting model from section~\ref{sec.exp.fore1} to \ref{sec.exp.foreend}.
In section \ref{sec.exp.fore1} we evaluate our action sequence forecasting model with cross-entropy loss.
Then in section \ref{sec.loss.eval} we evaluate the impact of new loss functions for action sequence forecasting.
After that in section ~\ref{sec.next.action} we compare our model with baselines models for predicting the next action.
We also evaluate the performance of our model when predicting the next action conditioned on the last three actions in section~\ref{sec.exp.foreend}.  
Finally, we evaluate weakly supervised action forecasting in section~\ref{sec.exp.wsaf} and compare with other published methods.
}

\begin{table*}[t]
	\caption{GRU Encoder-Decoder performance on action sequence forecasting}
	\label{tbl:base-seq-seq}
	\centering
		\begin{tabular}{|l|c|r|r|r|r|} \hline
			Dataset & Setup & BLEU-1 (\%) & BLEU-2 (\%) & Seq. Item. Acc (\%) & mAP \\ \hline
			Charades & Random  	& 1.04 & 0.35 	& 0.28 & 4.40 \\
			Charades & Classification 	& 15.26& 2.78	& 5.35 & 28.40 \\ 
			\R{Charades} & \R{Forecasting (Mean+GRU)} 		&  \R{5.15} & \R{1.87} 	& \R{2.30} & \R{5.90} \\
			\R{Charades} & \R{Forecasting (GRU-ED)} 		&  \R{5.75} & \R{2.18} 	& \R{1.53} & \R{5.10} \\			
			Charades & Forecasting (GRU-ED Att.) 	&  7.95 & 2.87 	& 2.60 & 6.10 \\ \hline		
			
			Breakfast & Random  		& 1.33 & 0.49 	& 0.70 & 7.53 \\
			Breakfast & Classification 	& 51.83 & 37.38	& 26.35   & 46.89  \\ 
			\R{Breakfast} & \R{Forecasting (Mean+GRU)}	& \R{25.65} & \R{10.23}	& \R{18.11}   & \R{24.94}  \\ 		
			Breakfast & Forecasting (GRU-ED Att.) 	& 34.56 & 21.15	& 21.29   & 30.24   \\ \hline
			
			MPII-Cooking & Random  	& 1.28 & 0.48 	& 0.47 & 6.53 \\
			MPII-Cooking & Classification 	& 25.74& 14.34 	& 14.86   &  20.60  \\ 
			MPII-Cooking & Forecasting (GRU-ED Att.) 	& 8.70 & 4.10 	&  4.50  &  10.80 \\  \hline
		\end{tabular}
\end{table*}

\subsection{How well does it perform in action sequence forecasting?}
\label{sec.exp.fore1}
In this section we evaluate our action sequence forecasting model using all three datasets.
During training, for each given video $X$ and the action sequence $Y= \left< y_1, y_2, \cdots, y_N \right>$, our model take feature sequence $X^o$ corresponding to observed action sequence $Y^o=\left< y^o_1, \cdots, y^o_i \right>$ and then predict future action sequence $Y^u=\left< y^u_{i+1}, \cdots, y^u_N \right>$ for all $i$ values (i.e. for $i=1,\cdots,N-1$). 
The observed $i^{th}$ action symbol is denoted by $ y^o_i$, and corresponds to a real action e.g. "opening fridge".
We use this action sequence sampling strategy to evaluate test videos for all possible $i$ values.
Unless otherwise specified, we use this strategy for training and testing which we call as the \textbf{Action Sequence Forecasting Setup}.
With this augmentation strategy, we obtain much larger dataset for training and evaluation.
This setup is different from what has been done in prior work\cite{AbuFarha2018}.

We report results using our GRU-based encoder-decoder model trained with attention (GRU-ED Att.) and traditional cross-entropy loss for action sequence forecasting.
As a baseline, we report results for random performance. 
In this case, for a given video, we randomly generate the next score vector to obtain the next action symbol for the unseen sequence. 
As the second baseline, we process the entire video feature sequence to obtain the full action sequence denoted by \textbf{Classification Setup}.
In this case, we observe the feature sequence $X^o$ corresponding to all actions and then output the action sequence $Y= \left< y_1, y_2, \cdots, y_N \right>$. 
We do this using the same model presented in section~\ref{sec.future.act.seq.gen} and cross-entropy loss. 
Results obtained by sequence classification model serves as a soft upper bound for the action sequence forecasting model.

\R{
\textbf{Additional baselines:}
To validate the effectiveness of attention mechanism, we also compare results with GRU encode-decoder without attention denoted by GRU-ED.
To validate the effectiveness of encode-decoder architecture, we also report results using another GRU baseline where the input to GRU is the mean I3D feature.
This model is denoted by Mean+GRU.
}

Results are shown in Table~\ref{tbl:base-seq-seq}. 
We make several observations.
First, our model performs significantly better than the random performance.
Sequence item classification accuracy (which is a strict measure) reflects the difficulty of action sequence forecasting task.
In the forecasting setup, we obtain item classification accuracy of 2.60, 4.50, and 21.29 where the random performance is 0.28, 0.47, and 0.70 on Charades, MPII cooking and Breakfast respectively. 
The random performance indicates the difficulty of forecasting task. Our model is 10-30 times better than random performance. 

The difference in results between classification and forecasting setups is not too drastic, especially for Breakfast and Charades.
Our classification model obtains seq. item accuracy of 5.35 while our forecasting model reach 2.60 on Charades.
Similarly for MPII cooking dataset, the classification model obtains 14.86 and our action forecasting model's performance is 4.50.
Interestingly, seq. item classification accuracy of 26.35 and 21.29 is obtained for classification and forecasting models respectively on Breakfast.
For action forecasting task, the Charades dataset is the most challenging and the least is Breakfast dataset.
Interestingly, for BLEU-2, the classification model obtains 2.78 while future action forecasting model performs better on Charades dataset.

\R{
The encoder-decoder model without attention (GRU-ED) improves results over Mean+GRU model on BLEU-1 and BLEU-2 scores on the challenging Charades dataset.
Interestingly, when attention mechanism is employed to the GRU encoder-decoder, the results improve over the Mean+GRU model on all metrics.
BLEU-1 score is improved from 5.15 to 7.95 and BLEU-2 score from 1.87 to 2.87.
Furthermore, sequence item classification accuracy is improved from 2.30 to 2.60.
These results suggest the effectiveness of encoder-decoder architecture with attention for action forecasting on the most challenging dataset.
On the Breakfast dataset, we see even massive improvements, where the GRU-ED Att. model obtains an improvement of 8.91 for BLEU-1 and 10.92 for BLEU-2.
Similarly, we see an improvement of 3.18\% for sequence item classification accuracy.
We conclude that GRU encoder-decoder with attention is effective for future action sequence forecasting problem.
}
These results indicate the effectiveness of our method for future action sequence forecasting task.
However, these results also suggest that there is more to do. Later in the experiments, we show how to improve these results.

\subsection{What is the impact of loss functions?}
\label{sec.loss.eval}

In this section we evaluate our method using the uncertainty and optimal transport loss functions for action sequence forecasting setup. 
The uncertainty loss consist of two parts in Equation~\ref{eq.un}.
\begin{enumerate}
	\item the effect of the fraction of past observations ($1-exp(\mathcal{P}/\mathcal{N})$) denoted by \emph{$L_{un}$-past-only}.
	\item the extent of future predictions ($exp(-q)$) denoted by \emph{$L_{un}$-future-only}. 
\end{enumerate}
First, we analyze the impact of these two terms separately and then evaluate them jointly. 
We also demonstrate the impact of optimal transport loss alone ($L_{ot}$).
Finally, we evaluate combination of all losses where we set the $\beta$ of Equation~\ref{eq.ot} to be 0.001. 
Results are reported in Table~\ref{tbl:losses}.
\begin{table}[t]
	\caption{Evaluating the impact of uncertainty losses and the optimal transport loss. Acc. is the sequence item classification accuracy.}\smallskip
	\label{tbl:losses}
	\centering
	
	\begin{tabular}{l|r|r|r|r} \hline
		Loss & BLEU-1 & BLEU-2 &  Acc. (\%) &  mAP (\%)   \\  \hline
		\multicolumn{5}{c}{Charade dataset.}\\ \hline
		Cross-entropy 			& 7.95 & 2.87 & 2.6 & 6.1\\
		$L_{un}$-past-only 		& 8.11 & 2.98& 2.6 & 6.1  \\
		$L_{un}$-future-only 	 & 8.61 & 3.11& 2.8 & 6.4 \\
		$L_{un}$-{both} 		 & 8.80 & 3.30& 2.9 & 7.2 \\
		$L_{ot}$ 				 & 7.73 & 3.06 & 3.4 & 7.2\\
		$L_{ot}$ + $L_{un}$-future-only	 & \textbf{9.59} & \textbf{3.92} & \textbf{4.0} & \textbf{8.2} \\ \hline
		\multicolumn{5}{c}{MPII Cooking dataset.}\\ \hline
		Cross-entropy 			& 8.70 & 4.10 & 4.50 & 10.80\\
		$L_{un}$-future-only	& 9.22 & 5.00 & 5.64 & 10.36\\ 
		$L_{ot}$				& 8.20 & 4.75 & 6.15 & 11.30\\ 
		$L_{ot}$ + $L_{un}$-future-only	& \textbf{11.43} & \textbf{6.74} &  \textbf{8.88} & \textbf{12.04} \\ \hline		
	\end{tabular}
\end{table}

From the results in Table~\ref{tbl:losses}, we see that both uncertainty and optimal transport losses are more effective than the cross-entropy loss which justifies our hypothesis about these new loss functions.
Interestingly, the loss term($L_{un}$-future-only), obtains the best results for BLEU scores while OT loss obtain best action sequence classification accuracy for an individual loss.
The combination of two uncertainty losses perform better than individual ones.
Combination of both $L_{ot}$ and $L_{un}$-future-only perform much better than all others obtaining a significant improvement in BLEU-1 and BLEU-2 scores from best of 7.95 to 9.59 and 2.87 to 3.92 on Charades dataset.
Similar trend can be seen for MPII-Cooking with consistent improvements.
This shows that optimal transport loss and $L_{un}$-future-only are complimentary to each other.
Though two uncertainty losses perform better than cross-entropy loss, unfortunately, the combination of all three losses do not seem to be useful. 
Perhaps we need a better way to combine both uncertainty losses with the OT loss which we leave for further investigation in the future.

We visualize some of the obtained results in figure~\ref{fig.results1}.
Interestingly, our method is able to generate quite interesting future action sequences. In the first example, our method accurately obtain four out of five actions. In the second example, it predicts two actions correctly, however the predicted action sequence seems plausible though it is not correct.


\subsection{How does it work for predicting the next action?}
\label{sec.next.action}.

In this section, we evaluate the impact of our sequence-to-sequence encoder-decoder architecture for predicting the next action.
For a given observed sequence $Y^o=\left< y^o_1, \cdots, y^o_i \right>$, the objective is to predict the next action $y^u_{i+1}$ for all  $i$ values of the video. Once again $y^o_i$ is the i-th action of the video.
As before, we generate all train and test action sequences.
For comparison, we also use two layered fully connected neural network (MLP) which applies mean pooling over the observed features and then use MLP as the classifier. 
Similarly, we also compare with a standard LSTM which takes the input feature sequence and then predict the next action only.
For our method and two baselines (LSTM, MLP), we use the same hidden size of 512 dimensions.
For all models, we use the same activation function, i.e. \texttt{tanh()}.
We report results in Table~\ref{tbl:next-action}.
\begin{table}[t]
\caption{Performance comparison for predicting next action. MLP is the multiple layered perceptron.}\smallskip
\label{tbl:next-action}
\centering
\begin{tabular}{|l|c|c|c|c|c|c|} \hline
& \multicolumn{2}{c}{Charades} & \multicolumn{2}{|c}{MPII Cooking} & \multicolumn{2}{|c|}{Breakfast} \\ \cline{2-7}
Method & Acc. (\%) & mAP & Acc. (\%) & mAP &  Acc. (\%) & mAP \\  \hline
MLP & 3.9 & 1.7 & 7.1 & 4.1  & 16.2  & 8.8 \\
LSTM & 2.5 & 1.3 & 2.4 & 3.0 & 4.3 & 3.0\\ 
Our & \textbf{6.8} & \textbf{3.0} & \textbf{11.0} & \textbf{9.2} & \textbf{16.4} &  \textbf{11.8}\\ \hline
\end{tabular}
\end{table}

First, we see that MLP obtains better results than LSTM.
Second, our sequence-to-sequence method with attention performs better than both LSTM and MLP methods.
MLP obtains 1.7 mAP for predicting the next action indicating features do not contain enough information about future and more complicated mechanism is need to correlate past features with the future action.
Our method obtains far better results than these two baselines indicating the effectiveness of our sequence-to-sequence architecture for next action prediction task.
We conclude our model is better suited for future action prediction than MLP and LSTM.


\subsection{What if we only rely on three previous actions?}
\label{sec.exp.foreend}

In this experiment we evaluate the performance of our model when we predict the next action using only the three previous actions. 
Here we train and test our method using all augmented action sequences. As before we use I3D features from the seen three actions and aims to predict the next action class. 
We also compare traditional cross-entropy with ($L_{ot}$ + Cross-entropy) loss. 
Results are reported on Table~\ref{tbl:past3}.
\begin{table}[t]
\caption{Action forecasting performance for using only the features from previous three actions on Charades dataset.}\smallskip
\label{tbl:past3}
\centering

\smallskip\begin{tabular}{l|c|c} \hline
Loss & Accuracy (\%) &  mAP (\%)  \\  \hline
Cross-entropy 			& 3.54 & 1.7    \\
$L_{ot}$ + Cross-entropy	& 6.25 & 2.3 \\ \hline
\end{tabular}
\end{table}

\begin{table*}[t]
	\caption{Comparison of action forecasting methods using Breakfast dataset only using features. 
	The best results using Fisher Vector (FV) features are \U{underlined}. Overall best results are shown in bold. 
	All results from \cite{AbuFarha2018} and \cite{ke2019time} use frame level annotations and therefore fully supervised. Cases where our weakly supervised method outperforms prior supervised state-of-the-art are underlined with red colour.}\smallskip
	\label{tbl:pq}
		\centering
		\smallskip\begin{tabular}{|l|c|c|c|c|c|c|c|c|} \hline
observation (\%) & \multicolumn{4}{|c|}{20\%} & \multicolumn{4}{|c|}{30\%} \\ \hline
prediction  (\%) & 10\% & 20\% & 30\% & 50\%  & 10\% & 20\% & 30\% & 50\% \\ \hline
Grammar\cite{AbuFarha2018}    		& 16.60   & 14.95   & 13.47   & 13.42   & 21.10   & 18.18  & 17.46  & 16.30 \\ 
Nearest Neighbor\cite{AbuFarha2018} & 16.42   & 15.01   & 14.47   & 13.29   & 19.88   & 18.64  & 17.97  & 16.57 \\
RNN\cite{AbuFarha2018} 				& 18.11   & 17.20   & 15.94   & 15.81   & 21.64   & 20.02  & \U{19.73}  & {19.21} \\ 
CNN\cite{AbuFarha2018}				& 17.90   & 16.35   & 15.37   & 14.54   & 22.44   & 20.12  & 19.69  & 18.76 \\ 
Time-Condition~\cite{ke2019time}    & 18.41   & 17.21   & 16.42   & 15.84   & 22.75   & 20.44& 19.64 & \U{19.75} \\
Our fully supervised - FV  		    &\U{18.75}&\U{18.35}&\U{17.78}&\U{17.00}&\U{23.98}&\U{21.93}& 19.70  & 19.58 \\ 
Our fully supervised - I3D& \B{23.03}	 & \B{22.28}  & \B{22.00} &\B{20.85}&\B{26.50}& \B{25.00}  & \B{24.08}  & \B{23.61} \\ \hline 
			  		
Our weakly supervised - FV& \I{18.60}  & 16.73  & 14.80  & 14.65  & \I{23.80}  & \I{21.15}  & 19.25  & 16.83  \\ 

Our weakly supervised - I3D& 21.70  & 18.85  & 16.65  & 14.58  & 26.20  & 21.78  & 20.43  & 16.50 \\ \hline
		\end{tabular}
\end{table*}

\begin{table*}[t]
	\centering
	
	\R{\caption{Comparison of action forecasting methods using 50Salads dataset only using features. 
		We report results using Fisher Vector (FV) features used in previous methods. 
		Overall best results are shown in bold. 
		When our weakly supervised method outperforms prior methods, it is underlined.
		All results from \cite{AbuFarha2018} and \cite{ke2019time} use frame level annotations and therefore fully supervised.}
	\smallskip
	\label{tbl:pq.50salad}
		\smallskip
		\begin{tabular}{|l|c|c|c|c|c|c|c|c|} \hline
			observation (\%) & \multicolumn{4}{|c|}{20\%} & \multicolumn{4}{|c|}{30\%} \\ \hline
			prediction  (\%) & 10\% & 20\% & 30\% & 50\%  & 10\% & 20\% & 30\% & 50\% \\ \hline
			Grammar\cite{AbuFarha2018}    			& 24.73 & 22.34 & 19.76 & 12.74 & 29.65 & 19.18 & 15.17 & 13.14  \\ 
			Nearest Neighbor\cite{AbuFarha2018}    	& 19.04 & 16.1  & 14.13 & 10.37 & 21.63 & 15.48 & 13.47 & 13.90  \\
			RNN\cite{AbuFarha2018} 					& 30.06 & 25.43 & 18.74 & 13.49 & 30.77 & 17.19 & 14.79 & 9.77	 \\ 
			CNN\cite{AbuFarha2018}				    & 21.24 & 19.03 & 15.98 & 9.87  & 29.14 & 20.14 & 17.46 & 10.86 \\ 
			Time-Condition~\cite{ke2019time} 		& 32.51 & 27.61 & 21.26 & 15.99 & 35.12 & 27.05 & 22.05 & 15.59 \\
Our fully supervised - FV& \B{39.32} & \B{31.39} & \B{27.01} & \B{23.88} & \B{41.73} & \B{32.73} & \B{31.44} & \B{26.39} \\ \hline		
Our weakly supervised - FV 				& \U{36.41} & 26.33 & \U{23.40} & 15.45 & \U{35.38} & 26.37 & \U{23.74} & \U{18.44} \\	\hline
		\end{tabular}}
\end{table*}

First, even for our method, we see a drop in performance from the results 
reported in previous experiment in Table~\ref{tbl:next-action}.
When we predict the next action using all previous action features, with the 
cross-entropy loss, we obtain a classification accuracy of 6.8\% in 
Table~\ref{tbl:next-action} whereas, in Table~\ref{tbl:past3}, our cross-entropy 
method obtains 3.54\% only.
This suggests that it is better to make use of all available information from 
observed video features and just let the attention mechanism to find the best 
features.
Secondly, the optimal transport loss combined with cross-entropy loss 
improve results indicating it is complimentary even in this 
constrained case.
For this experiment there is no need to make use of uncertainty loss as there 
is only one action to predict.\\

\subsection{Evaluating weakly supervised action forecasting and comparison to other SOA methods.}
\label{sec.exp.wsaf}
In this section we evaluate our weakly supervised action forecasting model presented in section~\ref{sec:wsaf}.
For this experiment we use Breakfast dataset \R{and commonly used 50Salads dataset~\cite{Stein2013}}.
In all prior experiments, we focus on forecasting future action sequence whereas most recent methods in the literature take a somewhat different approach\cite{AbuFarha2018,Gammulle2019,ke2019time}. 
These methods observe $p$\% of the video and aims to predict future actions for $q$\% of the video assuming length of video is known and frame level action annotations (at least the start and end of each action is known) are provided.
In this section we follow the protocol used in\cite{AbuFarha2018,Gammulle2019,ke2019time}.
However, our weakly supervised method does not make use of any frame level annotations during training.

First, we compare our fully supervised method (section~\ref{sec:saf}) against the weakly supervised method using I3D features on Breakfast dataset.
We compare our results with\cite{AbuFarha2018,ke2019time} and report mean per class accuracy as done in\cite{AbuFarha2018,ke2019time}. 
Unfortunately, the method in \cite{Gammulle2019} uses ground truth action sequence labels (the observed actions) during inference which is not a realistic setup.
As a fair comparison with methods proposed in\cite{AbuFarha2018,ke2019time}, we also experiment with the Fisher Vector (FV) features used in\cite{AbuFarha2018,ke2019time}. 
Results for Breakfast dataset are reported in Table~\ref{tbl:pq}.

When we use I3D features, our supervised method outperforms all baselines presented in \cite{AbuFarha2018,ke2019time} by a large margin, including larger prediction percentages such as 0.5. 
On average, our  supervised method obtains an improvement of 4.6\% over the best prior model in \cite{ke2019time}. 
Specifically,  the biggest average improvement is obtained when we observe only the 20\% of video. 
In this case, the average improvement is 5.1\% across all prediction percentages.
We also see a consistent improvement over all (p\%) percentages. 

When we use I3D features, our \emph{weakly supervised} method also outperforms fully supervised results of \cite{AbuFarha2018,ke2019time} in majority cases.
It fails only in two extreme cases, e.g., when predicting 50\% into the future.
This indicates the power of I3D features and the effectiveness of our weakly supervised method.
With I3D, our weakly supervised method is only 3.8\% behind our fully supervised method on average
and in one instance it is only 0.3\% behind fully supervised results (i.e. observe 30\% and predict 10\%).
As our weakly supervised method does not use any frame level annotations, this is a positively surprising result.
Even more conclusive trend can be seen when we use Fisher Vector features.

Most interestingly, our weakly supervised results are comparable to \cite{AbuFarha2018,ke2019time} when we use Fisher Vector features (FV).
Somewhat surprisingly, when predicting 10\% to the future (p=10\%), our weakly supervised method obtains better results than supervised methods of \cite{AbuFarha2018,ke2019time} indicating the effectiveness of our weakly supervised model presented in section~\ref{sec:wsaf}.
Our weakly supervised method is only 1.4\%, 0.6\% behind our supervised and recent~\cite{ke2019time} methods respectively.
Most interestingly, it is 0.1\% better than supervised CNN method of~\cite{AbuFarha2018}. 
Our weakly supervised method performs relatively better when we observe more data (i.e. results for 30\% observation is comparable to supervised performance of \cite{AbuFarha2018,ke2019time}).
In three out of eight cases, our weakly supervised method (with FV) outperforms supervised state-of-the-art methods such as \cite{AbuFarha2018,ke2019time}.
We attribute this improvement to the model architecture and the attention mechanism. 

Specifically, the use of bidirectional GRU helped to improve results. Bi-directional encoding allows us to better exploit temporal dependencies in feature sequence.
Furthermore, in our implementation we make sure that the feature dimensions of $X,H^o,H^u$ are the same --see section~\ref{sec:wsaf}.
We notice that lower or higher dimensions for $H^o$ and $H^u$ hinder the performance.
One interesting question is weather one should enforce the distribution of future hidden features $P(H^u)$ correlate with unseen future features distribution $P(X^u)$? We leave this question as a future exploration.

We also notice that the loss function in Eq.~\ref{eq.loss.joint} plays a special role. 
Specifically, the best results are obtained when we give slightly higher importance to the second term of this loss by setting $\gamma$ to be 2.0.
However, very large $\gamma$ values (such as  $\gamma=5.0$ or $\gamma=10.0$) are worse than smaller values (e.g. $\gamma=1.0$).
Interestingly, ignoring the first part of the loss term also leads to poor performance.
This shows that the models' ability to obtain a good representation for observed and unobserved future frames is important when forecasting future actions.
This also somewhat confirms our hypothesis on mental time travel discussed in the introduction.
Specifically, we assume that humans correlate prior experiences and examples with the current scenario to perform mental time travel.
In this regard, it seems a better understanding of the past events perhaps help to improve future predictions.

\begin{figure}[t]
	\centering
	gt:\includegraphics[width=.20\columnwidth,height=2mm]{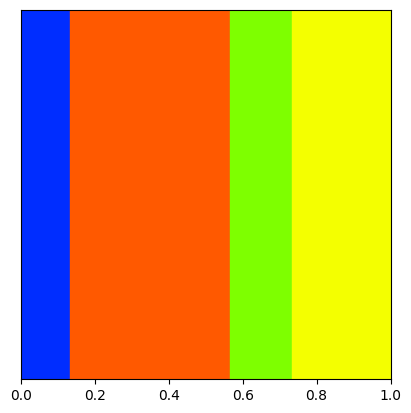}	
	\includegraphics[width=.20\columnwidth,height=2mm]{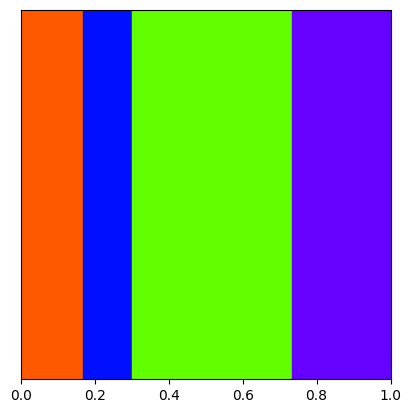}
	\includegraphics[width=.20\columnwidth,height=2mm]{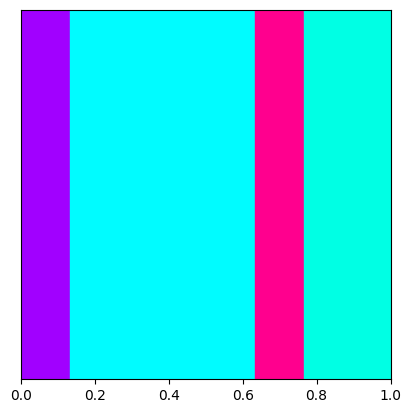}
	\includegraphics[width=.20\columnwidth,height=2mm]{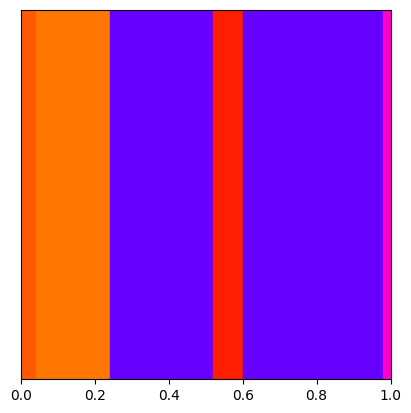}
	pr:\includegraphics[width=.20\columnwidth,height=2mm]{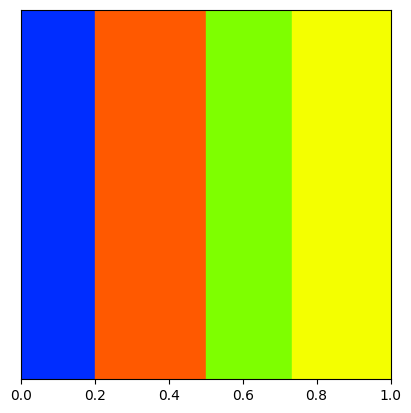}
	\includegraphics[width=.20\columnwidth,height=2mm]{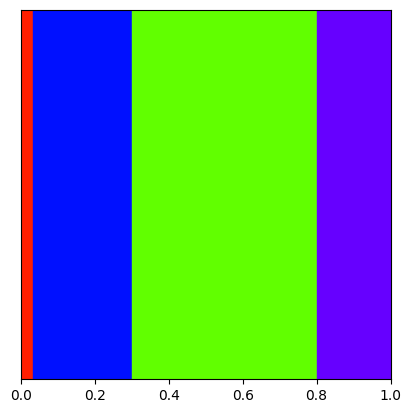}
	\includegraphics[width=.20\columnwidth,height=2mm]{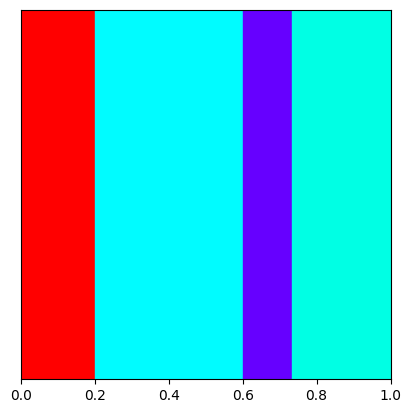}	
	\includegraphics[width=.20\columnwidth,height=2mm]{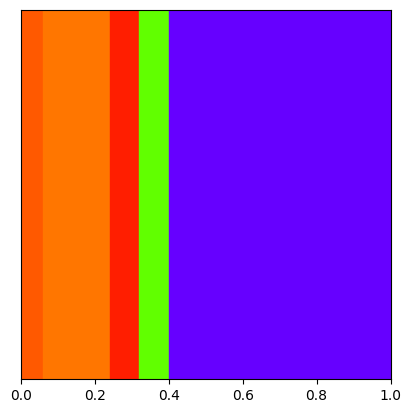}

	\caption{Illustration of ground truth and forecasted actions for some random videos. Each color represents an action.}
	\label{fig.vis.results}
\end{figure}

Visual illustration of some predictions are shown in figure~\ref{fig.vis.results}.
Interestingly, most of the time our method is able to get the action class correctly, although the temporal extent is not precise. 
Furthermore, there is significant smoothness in the prediction that we believe is due to the sequential learning used in our method.

\R{
Similarly, now we provide results on 50Salads dataset using the five fold cross validation with provided splits in~\cite{Stein2013} which is the protocol used in~\cite{AbuFarha2018,ke2019time}.
50Salads dataset has 50 videos with 17 fine-grained action classes where average length of a video is 6.4 minutes and contain 20 action instances per video on average.
We report the accuracy of predicted frames as mean over classes (MoC) and use the same Fisher vector features used in prior methods~\cite{AbuFarha2018}.
Results are reported in Table~\ref{tbl:pq.50salad}.
We see a significant improvement in results on this dataset compared to the improvements seen in the Breakfast dataset.
The recent Time-Condition~\cite{ke2019time} method obtains an average improvement of 4.62\% over prior RNN model of \cite{AbuFarha2018}.
Interestingly, our fully supervised method obtains a significant average improvement of 7.09\% over the Time-Condition~\cite{ke2019time} model.
Our weakly supervised method is 6.05\% lower than our supervised model, yet obtains better results than supervised Time-Condition~\cite{ke2019time} method by obtaining an average improvement of 1.04\% .
Compared to the breakfast dataset, the improvement in the 50Salads dataset is positively surprising.
By visual inspection of the data, we also notice that there is high temporal correlation in 50 Salads dataset, which might positively influence our model to generate accurate pseudo representation for future.
Once again, we attribute this improvement to the model architecture shown in figure~\ref{fig.weakarchi}, the effective use of attention mechanism, use of pseudo hidden states to represent future frame representation and effective use of new loss functions.  
}

\section{Conclusion.}
\label{sec.conc}
In this paper we presented a method to predict future action sequence from a partial observation of a video using 
a GRU-based encoder-decoder machine translation technique.
We showed the effectiveness of regularizing the cross-entropy loss for this task by catering the uncertainty of future predictions and 
the proposed optimal transport loss allowed us to further improve results.
We observed that conditioning on few past video frames is not sufficient to forecast future actions accurately.
It is better to make use of all available information and use attention mechanism to select the most relevant frames in the partially observed video. The attention mechanism helped our model to better exploit the context of activity and obtain accurate future predictions.

\R{
Weakly supervised action forecasting is an important problem and in this work we proposed an effective method by taking advantage of an architecture that is designed to correlate observed feature sequence with the future action sequence.
Our weakly supervised action forecasting model used an GRU encoder, three dedicated decoders and used an effective attention mechanism to obtain accurate actions for the future.
Using this attention method, our model predicted labels for future unseen frames at test time without using frame specific action labels during training.
It obtained competitive results compared to prior fully supervised methods and sometimes even outperformed them.
Our method is conceptually simple and potentially useful for many practical applications where one can train with easily obtainable coarse annotations of videos.
We believe our findings are insightful and useful for the development of future action forecasting methods.
}

\section*{Acknowledgment}
This research/project is supported by the National Research Foundation, Singapore under its AI Singapore Programme (AISG Award No: AISG-RP-2019-010). Any opinions, findings and conclusions or recommendations expressed in this material are those of the author(s) and do not reflect the views of National Research Foundation, Singapore.

\ifCLASSOPTIONcaptionsoff
  \newpage
\fi



%



\end{document}